\documentclass{ecai} 

\usepackage{latexsym}
\usepackage{amssymb}
\usepackage{amsmath}
\usepackage{amsthm}
\usepackage{booktabs}
\usepackage{enumitem}
\usepackage{graphicx}
\usepackage{colortbl}
\usepackage{xcolor}
\usepackage{multirow}  
\usepackage{makecell}  
\usepackage{hyperref}
\usepackage{subcaption}
\usepackage{bm}
\usepackage{algorithm}
\usepackage{algpseudocode}

\newcommand{\BibTeX}{B\kern-.05em{\sc i\kern-.025em b}\kern-.08em\TeX}

\begin{document}


\begin{frontmatter}


\paperid{123} 


\title{MAGIK: Mapping to Analogous Goals via Imagination-enabled Knowledge Transfer}


\author[A]{\fnms{Ajsal Shereef}~\snm{Palattuparambil}\thanks{Corresponding Author. Email: a.palattuparambil@deakin.edu.au}}
\author[B]{\fnms{Thommen George}~\snm{Karimpanal}}
\author[A]{\fnms{Santu}~\snm{Rana}} 

\address[A]{Applied Artificial Intelligence Initiative, Deakin University}
\address[B]{School of IT, Deakin University}
\address[A]{Applied Artificial Intelligence Initiative, Deakin University}


\begin{abstract}
Humans excel at analogical reasoning - applying knowledge from one task to a related one with minimal relearning. In contrast, reinforcement learning (RL) agents typically require extensive retraining even when new tasks share structural similarities with previously learned ones. 
In this work, we propose \emph{MAGIK}, a novel framework that enables RL agents to transfer knowledge to analogous tasks without interacting with the target environment. Our approach leverages an imagination mechanism to map entities in the target task to their analogues in the source domain, allowing the agent to reuse its original policy. Experiments on custom MiniGrid and MuJoCo tasks show that MAGIK achieves effective zero-shot transfer using only a small number of human-labelled examples. 
We compare our approach to related baselines and highlight how it offers a novel and effective mechanism for knowledge transfer via imagination-based analogy mapping.

\end{abstract}

\end{frontmatter}


\section{Introduction}
\label{sec:intro}
Deep Reinforcement Learning (RL) has achieved remarkable success in various domains, from robotic manipulation to autonomous navigation. However, a fundamental challenge remains—\textit{transferability} across related tasks and environments. Traditional Deep RL models often struggle to adapt to new but structurally related tasks, requiring extensive retraining. This limitation hinders the deployment of RL agents in real-world applications, where goals may frequently change \cite{zhang2018study}.

Traditional transfer learning in RL typically involves fine-tuning a pre-trained policy on a new task, adapting the agent’s behavior through additional interaction with the target environment. This approach assumes access to target-task data and requires further exploration and learning, which may be impractical in real-world scenarios where data collection is costly or risky. Zero-shot transfer—adapting to a new task without any additional training—remains especially challenging because it requires decoupling the agent’s knowledge of the environment from the specific goals it was trained to achieve. Most RL agents map perceived observations directly to a monolithic policy shaped by the reward structure. As a result, even subtle changes in the reward function or observation space can disrupt this mapping, making it difficult to adapt across tasks.

\begin{figure}[h]
    \begin{center}
        \includegraphics[width=\linewidth]{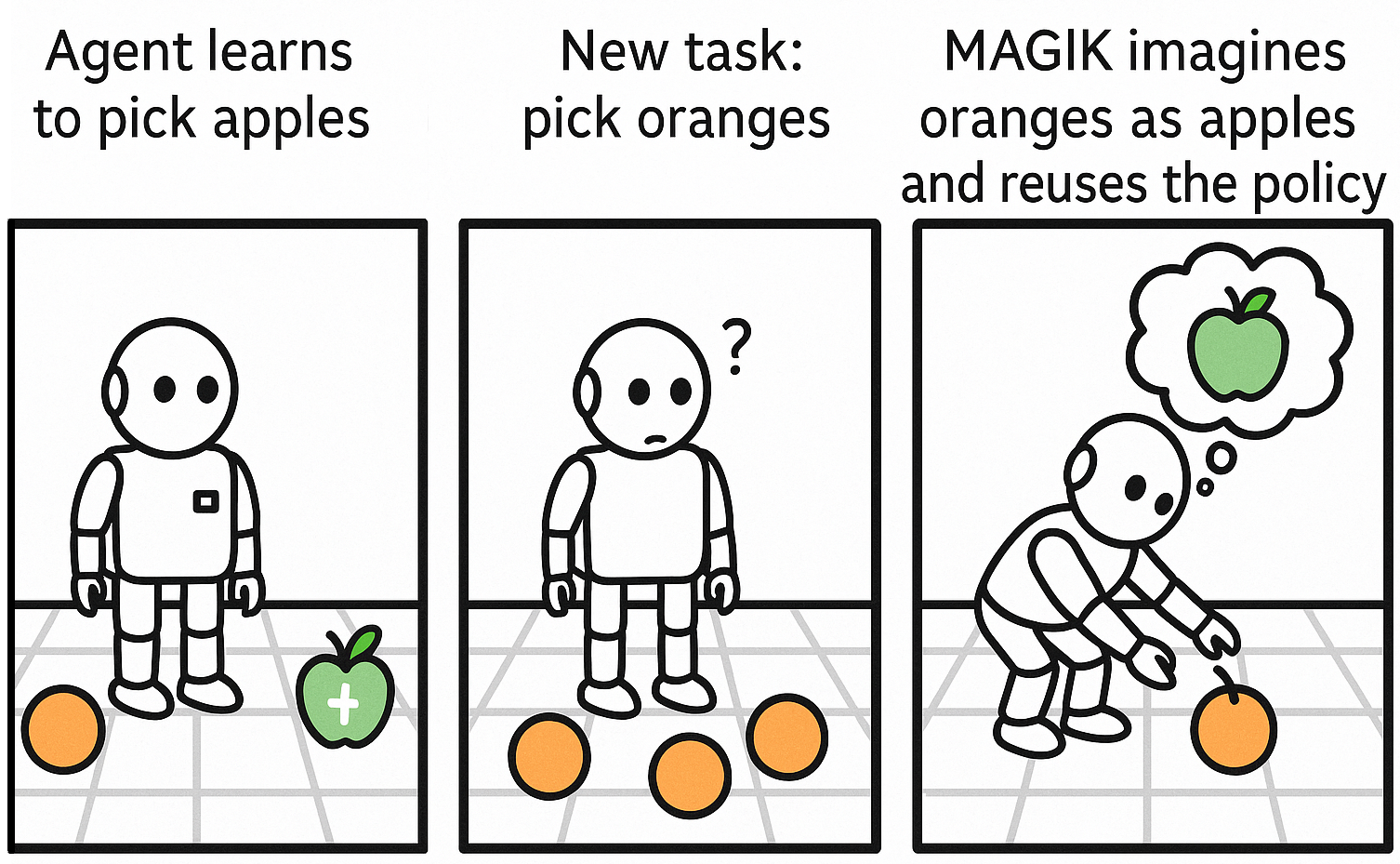}
    \end{center}
    \caption{The agent first learns a source policy for picking apples. Later, when instructed to pick oranges, it imagines the oranges as apples and reuses the same policy, effectively solving the new task without retraining.}
    \label{fig:idea}
    \vspace{10pt}
\end{figure}

To better understand the kind of transfer we are interested in, consider a simple example: an agent is trained in an environment containing both apples and oranges, but receives rewards only for picking apples. Later, the agent is instructed to perform a new task—picking oranges instead. A human in this situation would naturally think, “Why not pick the oranges the same way I picked the apples?”—reusing the learned skill by mentally substituting one target object for another, without needing to relearn from scratch. We aim to equip RL agents with this kind of cognitive flexibility. Specifically, we focus on tasks that are structurally similar but differ in goals or semantics. Rather than retraining, the agent is encouraged to “imagine” the oranges as apples and execute the same policy as if it were solving the original task. In essence, the agent solves the analogous task by mapping the current observation in the target task to a familiar one from the source task, thereby invoking its prior knowledge to act appropriately. Figure \ref{fig:idea} illustrate this idea.

To enable this kind of zero-shot transfer, we propose \textbf{M}apping to \textbf{A}nalogous \textbf{G}oals via \textbf{I}magination-enabled \textbf{K}nowledge (MAGIK), a generative approach based on a semi-supervised Variational Autoencoder (VAE). The core idea is simple: instead of learning a new policy for the target task, the agent imagines how the current situation would appear if it belonged to the original task it was trained on. By doing so, it can reuse its learned policy in a new context. This is enabled by a VAE that learns to disentangle task-agnostic aspects of an observation—such as the layout or object positions—from task-specific aspects—such as the identity of the rewarded object. Later, when faced with a new task, the agent generates (or imagines) a modified observation by combining the task-agnostic structure from the target task with task-specific information from the source task. This effectively maps the target observation into a familiar, source-like one where the agent already knows how to act.

While prior work on transfer in RL \citep{gamrian2019transfer,zhu2024distributional,zhang2017,touati2022does,chua2024learning} often requires fine-tuning or additional interaction with the target task, our method bypasses this entirely—requiring only a small number of labeled data points to help structure the latent space during training. MAGIK relies solely on what the agent has already learned and uses imagination to reinterpret the new task in familiar terms. This allows the agent to act effectively in new but related tasks without any further training. We empirically validate our approach in controlled RL environments, showing that MAGIK enables successful adaptation to new tasks with zero additional interaction.

Our contributions are as follows:
\begin{itemize}
    \item Proposal of a framework that achieves policy reuse in a zero-shot setting through imagination.
    \item We use a semi-supervised VAE architecture that disentangles class-agnostic and class-related features, allowing controlled state generation for analogous tasks.
\end{itemize}

The remainder of the paper is structured as follows. Section~\ref{sec:related_work} reviews related work on transfer learning and generative models in RL. Section~\ref{sec:p&b} outlines necessary background. Section~\ref{sec:problem_setup} discusses the problem setup.  Section~\ref{sec:method} presents our approach, including the architecture of the imagination network and the training procedure. Section~\ref{sec:experiments} provides experimental results, and Sections~\ref{sec:discussion_conclusion} offer discussion and concluding remarks.


\section{Related Work}
\label{sec:related_work}
Our work builds upon existing methods by integrating elements from transfer learning, imagination-based RL, semi-supervised VAEs, and structured latent representations.

\subsection{Transferability in Deep RL}  
A key strategy for improving transfer in RL involves learning shared latent representations that remain invariant across different tasks and domains \cite{higgins2017darla,cobbe2019quantifying,gamrian2019transfer}. These approaches enable agents to extract transferable features, allowing adaptation to new environments with minimal fine-tuning. However, many such methods rely on direct interaction with the target environment or assume the reward function remains unchanged across tasks. In contrast, our method facilitates transfer without direct exposure to the target domain, using a structured imagination model to infer representations aligned with analogous tasks. In \cite{shereef2024personalisation}, a personalised policy is constructed by dynamically integrating human intent via policy fusion.

Successor Features (SFs)~\cite{barreto2017successor} offer another promising direction for transfer by decomposing the value function into dynamics-dependent and reward-specific components. Generalised Policy Improvement (GPI)~\cite{barreto2018transfer} builds upon this to support zero-shot adaptation across tasks with shared dynamics and differing rewards. Recent work has scaled SFs to high-dimensional visual inputs using deep encoders \cite{borsa2018universal,chua2024learning,touati2022does,zhang2017}. The SF assume the linearity of the reward and this makes them limited in their applicability. Unlike SF, our approach does not assume linearity in the reward function and instead leverages structured imagination to manipulate the observation space, enabling transfer even when both rewards and observations change across tasks.

\subsection{Imagination and World Models in RL}  
Inspired by human cognitive abilities, several works have explored the idea of imagination in RL, where an agent learns an internal model to predict future states of the environment. Model-based RL approaches such as Dreamer \cite{hafner2019dream} and PlaNet \cite{hafner2018learning} train world models to improve sample efficiency and enable planning. These models rely on learned latent representations to generate rollouts, allowing agents to anticipate future states before acting.  

A related line of work focuses on \textit{imagination augmentation}, where an agent synthesises new experiences to enhance learning. Works such as Visual Reinforcement Learning with Imagined Goals \cite{nair2018visual} and Goal-Conditioned Imagination \cite{nasiriany2019planning} train models that generate imaginary states conditioned on desired goals. Our work aligns with this direction but differs in that we use an imagination network to generate analogous states in a transfer learning context, enabling an agent to infer a policy for an unseen but related task. In \cite{senadeera2023emote}, imagination net is used to infer the action value of the other agent in a multi-agent RL setting.

\subsection{Variational Autoencoders for Representation Learning in RL}  
Variational Autoencoders (VAEs) \cite{kingma2013auto,rezende2014stochastic} have been widely used for representation learning in RL, particularly for disentangling factors of variation in high-dimensional observations \cite{higgins2017beta,burgess2018understanding}. By learning structured latent spaces, VAEs can facilitate transfer by enabling the reuse of learned features across tasks \cite{higgins2017darla,ha2018world}.  

Our approach builds on this by incorporating a semi-supervised VAE with disentangled latents to separate class-agnostic and class-related features. This design enables us to modify the class-specific component of the latent space, allowing the agent to generate imagined states corresponding to analogous tasks. Additionally, we use Hilbert-Schmidt Independence Criterion (HSIC) \cite{gretton2005kernel} as a regularisation mechanism to enforce stronger independence between the latent variables, ensuring robust disentanglement and improving generalisation across tasks.  

\subsection{Semi-Supervised Learning for Structured Representations}  
Semi-supervised learning (SSL) provides a framework for learning from both labelled and unlabelled data, which has been extensively studied in the context of VAEs \cite{kingma2014semi,siddharth2017learning,hajimiri2021semi}. In RL, SSL has been used to improve performance, particularly when labelled supervision is scarce \cite{finn2016generalizing,zheng2023semi}. Our approach incorporates SSL to guide the structuring of the latent space, where a small fraction of labelled observations directs the latent representations toward meaningful partitions. This enables the effective generation of imagined states that align with analogous tasks, supporting policy transfer without explicit retraining on the target environment.

\section{Preliminaries and Background}  
\label{sec:p&b}
This section provides an overview of key concepts used in our approach, We used a Variational Auto Encoder for imagination net and Soft actor-critic (SAC) for policy learning. 

\subsection{Reinforcement Learning}
RL enables the agent to make decisions by interacting with an environment modelled as a Markov Decision Process (MDP). An MDP is defined by a tuple \(\langle\mathcal{S}, \mathcal{A}, T, R, \gamma\rangle\), where \( \mathcal{S} \) is the state space, \( \mathcal{A}\) is the action space, \( R: S \times A \to \mathbb{R} \) is the reward function, \( T: S \times A \to S \) is the transition function, and \( \gamma \in [0,1] \) is the discount factor. Upon each interaction, the agent observes a state \(s \in \mathcal{S}\), selects an action \(a \in \mathcal{A}\), receives a reward \(R(s, a)\), and transitions to a new state \(s' \sim P(\cdot \mid s, a)\). The agent's goal is to learn a policy \(\pi(a \mid s)\) that maximises the expected cumulative discounted return \(\mathbb{E}\left[ \sum_{t=0}^\infty \gamma^t r(s_t, a_t) \right]\).

\subsection{Variational Autoencoders (VAE)}  

VAEs \cite{kingma2013auto,rezende2014stochastic} are probabilistic generative models that learn structured latent representations from input data. A VAE consists of an encoder \(q_{\phi}(z | x)\), which maps inputs \(x\) to a latent distribution over the latent \(z\), and a decoder \(p_{\theta}(x | z)\), which reconstructs \(x\) by sampling latent variables. The model is trained by maximising the evidence lower bound (ELBO):  

\begin{equation}
\mathcal{L} = \mathbb{E}_{q_{\phi}(z|x)} [\log p_{\theta}(x | z)] - KL(q_{\phi}(z | x) || p(z)).
\end{equation}  

The first term forces reconstruction, while the Kullback-Leibler (KL) divergence acts as a regulariser to squish the posterior \(q_{\phi}(z | x)\) toward the prior distribution \(p(z)\).  

In our work, we extend the standard VAE to a semi-supervised VAE, where a fraction of data points is labelled to structure the latent space meaningfully. We provide the modified ELBO in Section \ref{sec:im_training}. The latent space is partitioned into task-relevant and task-agnostic components, enabling zero-shot transfer via latent-space transformations.  

\subsection{Soft Actor-Critic (SAC)}  

Soft Actor-Critic (SAC) \cite{haarnoja2018soft} is an off-policy deep RL algorithm based on the Maximum Entropy RL framework, which incentivise exploration by maximising both expected reward and entropy. Formally, SAC optimises the following objective:  

\begin{equation}
J(\pi) = \sum_t \mathbb{E}_{(s_t \sim \mathcal{T}, a_t \sim \pi}) \left[ r(s_t, a_t) + \alpha \mathcal{H}(\pi(\cdot | s_t)) \right],
\end{equation}  

where \(\mathcal{T}\) is the dynamics, \(\mathcal{H}(\pi(\cdot | s_t))\) is the entropy of the policy and \(\alpha\) controls the trade-off between exploration and exploitation. A SAC is composed of a policy network \(\pi_{\theta}(a | s)\) that learns a stochastic policy, Two Q-networks \(Q_{\phi_1}(s, a)\) and \(Q_{\phi_2}(s, a)\) to mitigate overestimation bias and a temperature parameter \(\alpha\), often learned to balance entropy and reward.  

SAC is well-suited for continuous action spaces and provides sample-efficient learning with robust exploration. In our work, SAC is used to train the agent in the source task, and the learned policy is transferred to the target task via our imagination-based transfer mechanism.

\section{Problem Setup}
\label{sec:problem_setup}

We consider a setting where an agent has learned a policy for a given source task and seeks to transfer that knowledge to an \textit{analogous} target task. Both tasks are situated in the same or similar environment, and we assume that while the reward function changes between tasks, the underlying structure of the environment remains largely consistent. Specifically, the observation space of the target task is assumed to be a subset of the source task's observation space. Crucially, the source and target tasks share certain \textit{transferable elements}—visually or semantically salient components of the observation (such as objects, object types, or spatial configurations) that are common across tasks and can support adaptation. These elements serve as anchors for adapting existing behaviors to new goals without retraining. Following prior work~\cite{higgins2017darla}, we assume that the source task contains all relevant transferable elements, while the target task may contain only a subset of them.

We assume that the agent has access to an offline dataset of observations, denoted as \( \mathcal{D} = \{s_{i}\}_{i=0}^{N} \), collected while learning the source task. A small subset of this dataset, \( \mathcal{D}_{l} = \{(s_{j}, c_j)\}_{j=0}^{M} \), is annotated with class labels \( c_j \), where \( s_j \in \mathcal{D} \). The number of labeled samples is limited.

In addition, we assume that each observation \(x \in \mathcal{D}\) can be explained by two latent variables that capture complementary factors of variation: a continuous variable \(z\), encoding task-agnostic features (e.g., background, layout), and a discrete variable \(c\), encoding task-specific information (e.g., goal-related object identity). These variables are designed to be \emph{mutually exclusive} (each accounts for a distinct aspect of the observation) and \emph{collectively exhaustive} (together they are sufficient to reconstruct \(x\)). This assumption enables structured representation learning, allowing the agent to selectively manipulate task-relevant features while preserving invariant structure across tasks.

Each task is modeled as a Markov Decision Process. Let \( \pi_{\theta}(a|s) \) be a policy parameterized by \( \theta \), which selects action \( a \in A \) from state \( s \in S^\mathcal{S} \) in the source task (denoted by superscript \( \mathcal{S} \)). Our goal is to infer an analogous policy \( \pi_{\theta}(a|\tilde{s}) \) for the target task, where \( \tilde{s} \in S^\mathcal{T} \) represents an observation in the target domain.

To enable this, we learn an imagination network \( \psi(\tilde{s}) \) such that \( \psi(\tilde{s}) = s' \in S^\mathcal{S} \)—a state aligned with the source domain but conditioned on the target task. This allows the agent to apply the source policy as if it were operating in the source environment. Thus, MAGIK introduces a generative mapping module \( \psi \) that transforms observations from the target domain into the source domain conditioned on the target task, enabling the agent to reuse its learned policy and induce behavior suitable for the new task.

\section{Methodology}
\label{sec:method}

Our goal is to develop a more generalised RL agent by enabling it to transfer its acquired knowledge to related tasks. To achieve this, we introduce an additional module—a semi-supervised Variational Autoencoder (VAE)\footnote{The same idea can also be achieved using GAN. Details are provided in the supplementary material, Section \ref{sec:gan_imagination}}—which is trained on the same dataset used for the agent's task learning. This VAE acts as a mapper from the target domain conditioned on the target task to the source domain. In other words, when the agent is deployed to perform the target task, the VAE maps the observation in the target to the source and executes the policy trained in the source task, thus enabling zero-shot transfer.

Formally, let \( \pi_{\theta}(a|s) \) be the policy learned in the source task. The analogous policy in the target task is obtained by a VAE parameterized by \( \psi \), which maps the target observation \( \tilde{s} \) to its source-aligned counterpart. This allows the agent to execute the policy as \( \pi_{\theta}(a|\psi(\tilde{s})) \).

Since the VAE operates in a semi-supervised manner, we assume that a small portion of the dataset is labeled by a human. We demonstrated, the labeled samples are randomly selected from the dataset. However, to further reduce the burden of human annotation, an active sampling strategy could be employed, where only the most informative or uncertain samples as determined by the classification head are queried for human labels. This would ensure that human supervision is concentrated on the most ambiguous cases, thereby improving label efficiency and scalability of the approach. These labeled samples play a crucial role in guiding the VAE to structure its latent space meaningfully. Labels correspond to different \textit{transferable elements} that define the task (e.g., apple vs. orange). Additional details on the labeling are provided in Section~\ref{sec:exp_setup}.

\subsection{Training of the Imagination Network}
\label{sec:im_training}

The Variational Autoencoder (VAE) is modified to incorporate two latent variables: $ z $, a continuous latent code sampled from a Gaussian distribution to capture task-agnostic features (e.g., background structure or spatial layout), and $ c $, a discrete latent code sampled using the Gumbel-Softmax trick \cite{jang2017categorical,maddison2017concrete} to represent task-specific factors (e.g., object identity or class-relevant information). These variables are designed to capture mutually exclusive and exhaustive aspects of the observation, with their disentanglement enforced using a small set of labelled observations and the Hilbert-Schmidt Independence Criterion (HSIC) as an auxiliary loss term.

During reconstruction, the decoder combines $ z $ and $ c $, inferred from the input observation, to reconstruct the original image.

The VAE is trained by maximizing the evidence lower bound (ELBO), which balances reconstruction accuracy and regularization for both labelled and unlabelled data. The overall objective sums the ELBO terms for these data types, enabling the model to learn robust, disentangled representations across supervised and unsupervised settings.

For labelled data, the ELBO is given by:
\begin{multline}
\log p_{\theta}(x,y) \geq 
\underbrace{\mathbb{E}_{q_{\phi}(z|x,y)} [\log p_{\theta}(x|z,y)]}_{\text{Reconstruction term}} + 
\underbrace{\mathbb{E}_{q_\phi(c|x)} [\log p(y|c)]}_{\text{Supervision term}} \\ 
- \underbrace{KL(q_{\phi}(z|x) \parallel p(z)) - KL(q_{\phi}(c|x) \parallel p(c))}_{\text{KL regularisation term}}.
\end{multline}

Here, the supervision term is crucial: since we typically define \(p(y \mid c)\) as a deterministic one-hot distribution, any probability mass assigned by \(q_{\phi}(c|x)\) to an incorrect class yields a large penalty in the ELBO. 
Equivalently, this reduces to maximising \(\log q_{\phi}(c=y \mid x)\), i.e., the log-likelihood of the correct class under the variational posterior. 
This makes the supervision term identical to the standard cross-entropy loss, ensuring that the discrete latent \(c\) is aligned with semantic class identity rather than arbitrary clusters. We refer the reader to the supplementary Section \ref{sec:elbo} for detailed explanation.

For unlabelled data, where supervision is not available, the ELBO simplifies to:
\begin{multline}
\log p_{\theta}(x) \geq 
\underbrace{\mathbb{E}_{q_{\phi}(z|x)} [\log p_{\theta}(x|z)]}_{\text{Reconstruction term}} \\ 
- \underbrace{KL(q_{\phi}(z|x) \parallel p_{\theta}(z)) - KL(q_{\phi}(c|x) \parallel p_{\theta}(c))}_{\text{KL regularisation term}}.
\end{multline}
Since labels are unavailable in this case, the supervision term is absent.

Additionally, to further enforce the disentanglement of \( z \) and \( c \), we incorporate the Hilbert-Schmidt Independence Criterion (HSIC) as an auxiliary loss term. 
HSIC measures the dependence between two random variables using kernel embeddings. 
Specifically, it is computed as:
\begin{eqnarray}
    \text{HSIC}(z, c) = \frac{1}{(n-1)^2} \text{Tr}(K_z K_c),
\end{eqnarray}
where \( K_z \) and \( K_c \) are the centered kernel matrices corresponding to \( z \) and \( c \), respectively. 
Here, \(\text{Tr}\) denotes the matrix trace and \(n\) is the number of samples. 
A lower HSIC value indicates reduced statistical dependence between \( z \) and \( c \), promoting better disentanglement. 
(The experimental evaluation is done by traversing both latent variables; results are provided in the supplementary materials, Section~\ref{sec:disentanglement}.)





The final objective function for training the VAE is given by:

\begin{eqnarray}
\mathcal{L} = -\text{ELBO}_{\text{labeled}} - \text{ELBO}_{\text{unlabeled}} + \lambda^{h} \cdot \text{HSIC}(\mathbf{z}, \mathbf{c}),
\end{eqnarray}

where $\lambda^{h}$ controls the strength of the independence regularisation.

The detailed derivation of the ELBO is provided in the supplementary section \ref{sec:elbo}.
\subsection{Imagination Network Architecture}
\label{sec:architecture}
The imagination network enables the agent to generate an imagined state by modifying class-related features while preserving class-agnostic information. The architecture consists of two parallel encoders, each with the same structure, designed to extract different aspects of the input observation.  

\begin{figure}[h]
    \begin{center}
        \includegraphics[width=\linewidth]{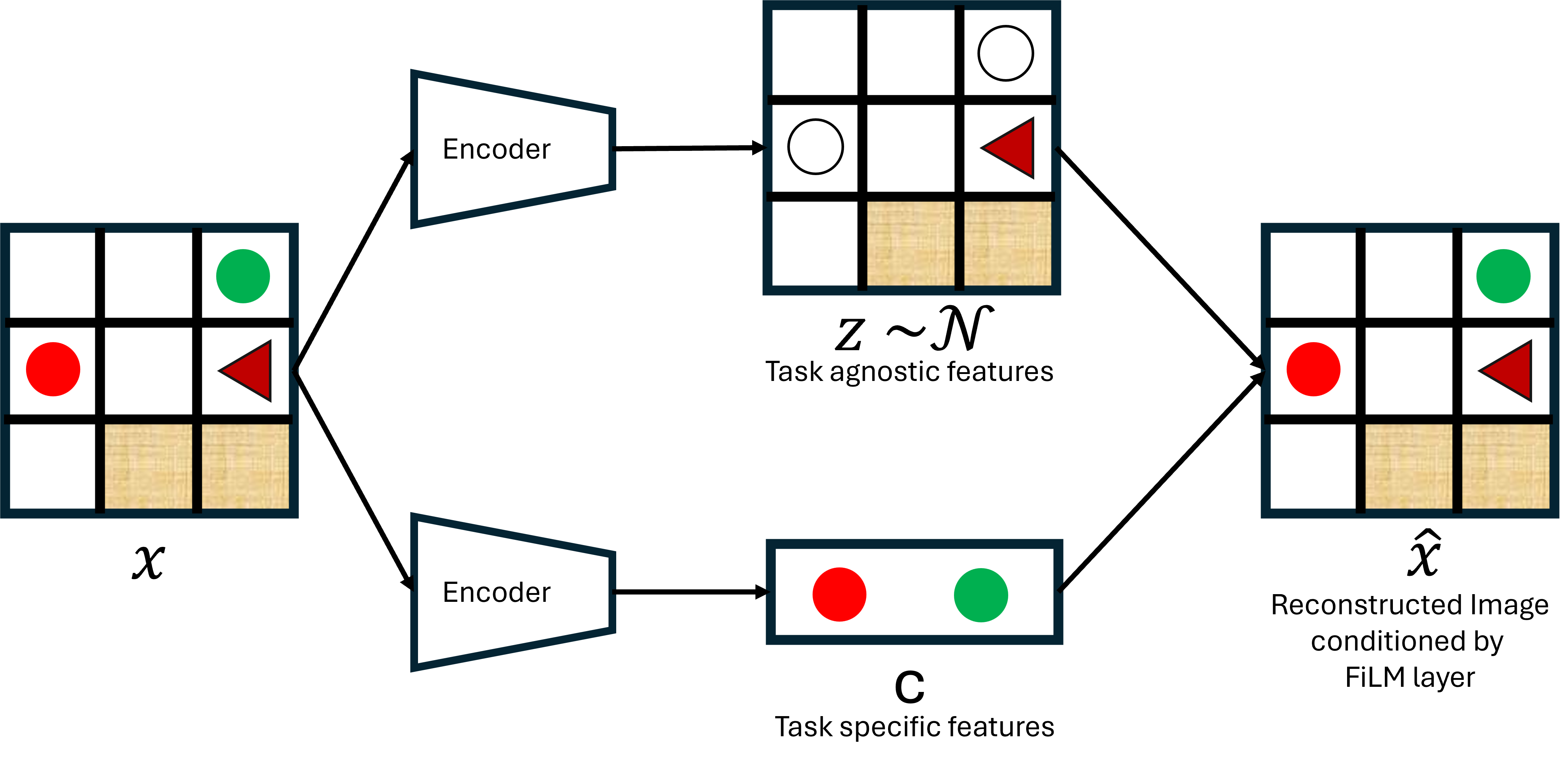}
    \end{center}
    \caption{Architecture of the VAE. The state observation (Both labelled and unlabelled) are fed to separate encoders to output class agnostic latent $z$ and class-related latent $c$. The decoder uses FiLM layers to condition the $c$ latent to $z$ and reconstruct the image.}
    \label{fig:vae_architecture}
    \vspace{15pt}
\end{figure}

The first encoder processes the input and passes its output through a bottleneck layer to produce a continuous latent variable \( z \), which captures class-agnostic features. The second encoder extracts class-related features and directly produces the discrete latent variable \( c \). This part of the network acts as the classification head. These two latent variables jointly serve as inputs to the decoder, which reconstructs the state. While reconstructing the image, we condition $c$ with $z$ using the FiLM architecture \cite{perez2018film}. A full architecture is given in Figure \ref{fig:vae_architecture} and for further details, please refer to supplementary Section \ref{sec:detailed_archtecture}.

Imagination is performed by altering the class label \( c \) while keeping the class-agnostic latent variable \( z \) unchanged.  This allows the network to generate an imagined state that retains the general structure of the observed state but aligns with the new class label. By leveraging this architecture, the agent can infer plausible state representations for analogous tasks without direct experience in the target setting. Algorithm \ref{alg:imagination_action} details how MAGIK selects action in the target environment.

\begin{algorithm}[H]
\caption{Zero-Shot Action Selection via Imagination in MAGIK}
\label{alg:imagination_action}
\begin{algorithmic}[1]
\Require Target observation $x^{\text{target}}$, source class $c^{\text{source}}$, encoder $q_\psi(z|x)$, decoder $p_\phi(x|z, c)$, pretrained source policy $\pi_{\text{source}}$
\Ensure Action $a^{\text{target}}$ to execute in the target environment
\vspace{1mm}
\State // \textbf{Encode Target Observation}
\State $z^{\text{target}} \sim q_\psi(z \mid x^{\text{target}})$  \Comment{Extract task-agnostic latent}
\vspace{1mm}
\State // \textbf{Imagine Source-Aligned Observation}
\State $\hat{x}^{\text{source-aligned}} \sim p_\phi(x \mid z^{\text{target}}, c^{\text{source}})$ \Comment{Reconstruct with source task identity}
\vspace{1mm}
\State // \textbf{Action Selection Using Source Policy}
\State $a^{\text{target}} \sim \pi_{\text{source}}(\hat{x}^{\text{source-aligned}})$
\vspace{1mm}
\State \Return $a^{\text{target}}$
\end{algorithmic}
\end{algorithm}

\section{Experiments}
\label{sec:experiments}

In this section, we empirically evaluate the following hypothesis:

\begin{itemize}
    \item \textbf{H1}: \textit{MAGIK} can successfully transfer knowledge to an analogous task in a target domain with sufficient transferable structure, even when the reward functions differ.
    \item \textbf{H2}: \textit{MAGIK} achieves comparable performance to a fine-tuned RL agent on a reward-modified task, while using significantly less additional data (Human labelling).
\end{itemize}

\subsection{Experimental Setup}
\label{sec:exp_setup}

To validate the effectiveness of our approach, we conduct experiments \footnote{The code is available \href{https://anonymous.4open.science/r/MAGIK-DC80/}{here}} in two environments: a custom image-based MiniGrid environment \cite{MinigridMiniworld23} and a feature-based MuJoCo Reacher environment \cite{towers2024gymnasium}. These environments were chosen to evaluate transfer performance under both visual and non-visual settings, with distinct action and observation spaces. To simulate real-world constraints on labeling, we simulated class labeling for training VAE as such no human were involved in labeling in our experiments.

\begin{figure}[htb]
    \centering
    \captionsetup[subfigure]{font=small, skip=0pt} 
    \captionsetup{skip=10pt} 
    \begin{subfigure}[b]{0.45\linewidth}
        \includegraphics[width=\linewidth]{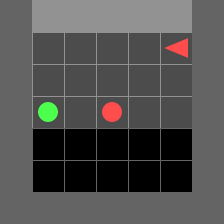}
        \caption{MiniGrid env.}
        \label{fig:minigrid}
    \end{subfigure}
    \hfill
    \begin{subfigure}[b]{0.45\linewidth}
        \includegraphics[width=\linewidth]{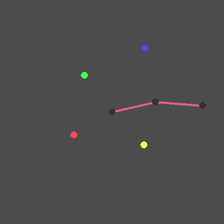}
        \caption{Reacher env.}
        \label{fig:reacher}
    \end{subfigure}
    \caption{Full view of the environments. The reacher is a feature-based observation. However, the snippet is a rendered image from the features}
    \label{fig:envs}
    \vspace{15pt}
\end{figure}

\textbf{MiniGrid (Discrete, Image-based)}:  
The MiniGrid environment presents a partially observable gridworld from an egocentric RGB perspective. The source task contains two objects—a green ball and a red ball. The agent is trained with a sparse reward signal, receiving a reward of +1 only when it picks the green ball. Since the red ball offers no reward, the agent learns to ignore it during source task training. Each episode has a maximum of 20 time steps, and the action space is discrete, consisting of original MiniGrid actions. See Figure \ref{fig:minigrid}.

Each observation is categorised into one of four classes based on the presence of the green or red ball in the agent’s field of view. The four classes used are detailed in Table~\ref{tab:vae_classes}. These labels guide the structured latent space in the imagination network and support transferability to analogous tasks.

\textbf{MuJoCo Reacher (Continuous, Feature-based):}  
The second environment is a modified version of the Reacher task in MuJoCo, where four coloured targets are randomly placed in each quadrant of the 2D space. In the source task, the agent learns to reach a specific target. In contrast, in the target task—introduced only at test time—the agent is required to reach a different target. The agent can only observe the target located in the quadrant where the reacher's tip currently resides.

The observation space is low-dimensional and includes the sine and cosine of the two joint angles, the angular velocities of the joints, the fingertip position, the target position, the target's colour, and the distance to the target. See Figure \ref{fig:reacher}.

Class labels in this environment are defined based on the colour of the target present in the same quadrant as the reacher's tip. These labels again serve to structure the latent space during training, enabling the imagination network to generate feature states corresponding to alternative tasks. The reacher is assumed to be reached if the fingertip is within a threshold distance to the target.

Overall, the experimental setup is designed to evaluate whether the proposed method can transfer across tasks with structurally similar but different goals with minimal data labeling.

\subsubsection{Evaluating \textbf{H1}: Transferability}
To evaluate \textbf{H1} in MiniGrid, we first train the agent using the SAC algorithm (Since the environment has discrete actions, we use discretised action selection in SAC) on the source task. We then transfer both the SAC agent and \textit{MAGIK} to three different target tasks:

\begin{enumerate}
    \item \textbf{Target 1}: Only the red ball is present in the target environment. The agent's task is to pick the red ball. This represents a scenario where the observation space is different as the target doesn't contain the green ball.
    \item \textbf{Target 2}: Both the red and green balls are present. The agent's task is to pick both balls.
    \item \textbf{Target 3}: Both the red and green balls are present, but the agent's task is to pick only the red ball. This represents a scenario where the reward function is significantly different from the source task.
\end{enumerate}

For Target 1 and Target 2, \textit{MAGIK} was instructed by tweaking the $c$ latent to imagine the target observation as belonging to the green ball class (Class 2) in the source environment whenever it classified the target observation as the red ball class (Class 1). \textit{MAGIK} is equipped with its own classification head (see architecture in Section~\ref{sec:architecture}). In all other cases, \textit{MAGIK} performed the same actions as the SAC agent.

For example, in Target 2, when the agent observes both the red and green balls (Class 3), \textit{MAGIK} first picks the green ball, as the SAC agent would. After the green ball is picked, the remaining observation contains only the red ball, which \textit{MAGIK} imagines as a green ball, thereby allowing it to pick the red ball efficiently.

For Target 3, \textit{MAGIK} applies a different imagination strategy:
\begin{itemize}
    \item Class 1 (only red ball) is imagined as Class 2 (only green ball).
    \item Class 2 (only green ball) is imagined as Class 4 (no ball).
    \item Class 3 (both red and green balls) is first imagined as Class 1 (only red ball) by removing the green ball from observation, and then further transformed to Class 2 (replacing the red ball with a green ball).
\end{itemize}
\begin{table}[t]
    \caption{Classification of observations for VAE training.}
    \centering
    \begin{tabular}{ll}
        \toprule
        {\bf Class} & {\bf Observation Content} \\
        \midrule
        Class 1 & Only red ball present \\ 
        Class 2 & Only green ball present \\ 
        Class 3 & Both red and green balls present \\ 
        Class 4 & Neither red nor green ball present \\
        \bottomrule
    \end{tabular}
    \label{tab:vae_classes}
\end{table}

We evaluated the zero-shot transfer performance of SAC and MAGIK to the target task, summarised in Table ~\ref{tab:transfer}. While both environments share structural similarities between objects and an analogous picking task, the SAC agent consistently failed to pick the red ball. Occasional successes with red balls or failures with green balls (observed in both agents) likely stem from stochastic policies and partial observability constraints, which similarly affect MAGIK as it employs the same SAC backbone. In contrast, MAGIK reliably accomplished the target tasks via imagination-based action execution, validating H1: The framework successfully transfers knowledge to analogous tasks in the target environment, demonstrating robust transferability. 

In the Reacher environment, the source task requires the agent to reach a blue target location, while the target tasks involve reaching other coloured targets—specifically red, green, and yellow. To enable transfer, the imagination network modifies observations from the target tasks by replacing the target colour with blue, effectively allowing the agent to interpret the new task through the lens of the source task. This substitution enables the learned policy to transfer without additional fine-tuning. The results of this transfer are summarised in Table~\ref{tab:transfer_reacher}\footnote{However, we found a performance drop if the source training data is not diverse enough. Details are provided in the Supplementary Section \ref{sec:add_exp}}.

\begin{table}[t]
\caption{Performance of agents in the MiniGrid environment, with SAC fine-tuned as described in Section~\ref{sec:data_efficiency}. Results are averaged over 5 seeds, each with 10 episodes, with standard deviation as the error. Red-shaded cells in the ``Green ball picked'' column for Target 3 indicate performance on a task not aligned with the target objective.}
\label{tab:transfer}
\centering
\begin{tabular}{llll}
\toprule
\multicolumn{1}{l}{\bf Target} & \multicolumn{1}{l}{\bf Agent} & \multicolumn{1}{l}{\bf \makecell{Green ball picked \\ (Out of 10)}} & \multicolumn{1}{l}{\bf \makecell{Red ball picked \\ (Out of 10)}} \\
\midrule
\multirow{7}{*}{Target 1} & SAC                                 & NA               & $2.00 \pm 0.58$ \\
                          & SAC-Fine tuned                      & NA               & $\bm{8.60 \pm 0.74}$ \\
                          & GAN \cite{gamrian2019transfer}      & NA               & $2.06 \pm 1.14$ \\
                          & SF-Simple \cite{chua2024learning}   & NA               & $8.40 \pm 0.89$ \\
                          & SF-Reconstruction \cite{zhang2017}  & NA               & $\bm{8.60 \pm 1.51}$ \\
                          & SF-Canonical \cite{kulkarni2016deep}& NA               & $0.00 \pm 0.00$ \\
                          & SF-Orthogonal \cite{touati2022does} & NA               & $0.40 \pm 0.54$ \\
                          & \textit{MAGIK} (Ours)               & NA               & $8.20 \pm 0.20$ \\
\midrule
\multirow{7}{*}{Target 2} & SAC                                 & $\bm{9.20 \pm 0.20}$  & $2.33 \pm 0.67$ \\
                          & SAC-Fine tuned                      & $8.20 \pm 0.37$  & $8.20 \pm 0.37$ \\
                          & SF-Simple \cite{chua2024learning}   & $2.40 \pm 0.54$  & $8.20 \pm 1.64$ \\
                          & SF-Reconstruction \cite{zhang2017}  & $4.40 \pm 0.89$  & $5.80 \pm 0.44$ \\
                          & SF-Canonical \cite{kulkarni2016deep} & $0.00 \pm 0.00$ & $0.00 \pm 0.00$ \\
                          & SF-Orthogonal \cite{touati2022does} & $0.20 \pm 0.44$  & $0.20 \pm 0.44$ \\
                          & \textit{MAGIK} (Ours)               & $\bm{9.20 \pm 0.37}$  & $\bm{9.40 \pm 0.24}$ \\
\midrule
\multirow{7}{*}{Target 3} & SAC                                 & \cellcolor{red!30}$8.80 \pm 0.37$  & $1.75 \pm 0.25$ \\
                          & SAC-Fine tuned                      & \cellcolor{red!30}$0.83 \pm 0.54$  & $\bm{9.00 \pm 0.54}$ \\
                          & SF-Simple \cite{chua2024learning}   & \cellcolor{red!30}$0.40 \pm 0.54$  & $7.80 \pm 1.64$ \\
                          & SF-Reconstruction \cite{zhang2017}  & \cellcolor{red!30}$0.80 \pm 0.83$  & $7.60 \pm 0.54$ \\
                          & SF-Canonical \cite{kulkarni2016deep}& \cellcolor{red!30}$0.00 \pm 0.00$  & $0.80 \pm 1.78$ \\
                          & SF-Orthogonal \cite{touati2022does} & \cellcolor{red!30}$0.20 \pm 0.44$  & $0.20 \pm 0.44$ \\
                          & \textit{MAGIK} (Ours)               & \cellcolor{red!30}$\bm{0.00 \pm 0.00}$  & $\bm{9.00 \pm 0.54}$ \\
\bottomrule
\end{tabular}
\end{table}

\begin{table}[htbp]
\caption{Performance of different agents in the reacher environment. SAC fine-tuned as in Section~\ref{sec:data_efficiency}. The data is averaged over 5 seeds, each with 20 episodes and standard deviation as the error.}
\label{tab:transfer_reacher}
\centering
\begin{tabular}{lll}
\toprule
\multicolumn{1}{l}{\bf Target} & \multicolumn{1}{l}{\bf Agent} & \multicolumn{1}{l}{\bf \makecell{Target reached \\ (Out of 20)}} \\
\midrule
\multirow{5}{*}{Red}      & SAC                               & $3.60 \pm 1.81$   \\
                          & SAC-Fine tuned for Red            & $\bm{19.80 \pm 0.44}$  \\
                          & SF-Simple \cite{chua2024learning}  & $3.00 \pm 3.90$   \\
                          & SF-Reconstruction \cite{zhang2017} & $5.60 \pm 2.30$   \\
                          & \textit{MAGIK} (Ours)             & $19.20 \pm 0.83$  \\
\midrule
\multirow{5}{*}{Green}    & SAC                               & $3.20 \pm 2.38$   \\
                          & SAC-Fine tuned for Green          & $\bm{19.40 \pm 0.89}$  \\
                          & SF-Simple \cite{chua2024learning}  & $2.00 \pm 1.41$   \\
                          & SF-Reconstruction \cite{zhang2017} & $4.60 \pm 4.15$   \\
                          & \textit{MAGIK} (Ours)             & $19.20 \pm 0.83$  \\
\midrule
\multirow{4}{*}{Blue}     & SAC                               & $19.80 \pm 0.44$  \\
                          & SF-Simple \cite{chua2024learning} & $2.40 \pm 2.19$   \\
                          & SF-Reconstruction \cite{zhang2017} & $4.60 \pm 3.28$   \\
                          & \textit{MAGIK} (Ours)             & $\bm{20.00 \pm 0.00}$  \\
\midrule
\multirow{5}{*}{Yellow}   & SAC                               & $3.80 \pm 1.92$   \\
                          & SAC-Fine tuned for Yellow         & $\bm{19.80 \pm 0.44}$  \\
                          & SF-Simple \cite{chua2024learning} & $0.80 \pm 1.09$   \\
                          & SF-Reconstruction \cite{zhang2017} & $4.60 \pm 4.15$   \\
                          & \textit{MAGIK} (Ours)             & $19.60 \pm 0.89$  \\
\bottomrule
\end{tabular}
\end{table}


\begin{figure}[htb]
    \centering
    \captionsetup[subfigure]{font=small, skip=5pt} 
    \captionsetup{skip=12pt} 
    \begin{subfigure}[b]{0.3\linewidth}
        \centering
        \includegraphics[width=\linewidth]{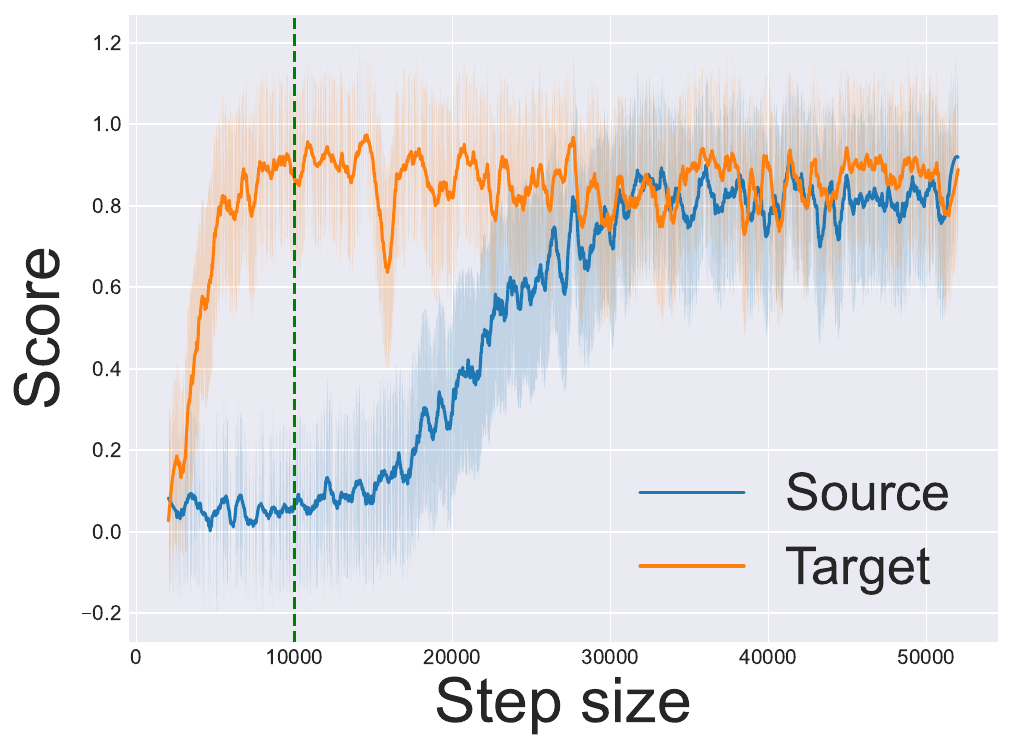}
        \caption{Fine-tuning in Target 1.}
        \label{fig:tr1}
    \end{subfigure}
    \hfill
    \begin{subfigure}[b]{0.3\linewidth}
        \centering
        \includegraphics[width=\linewidth]{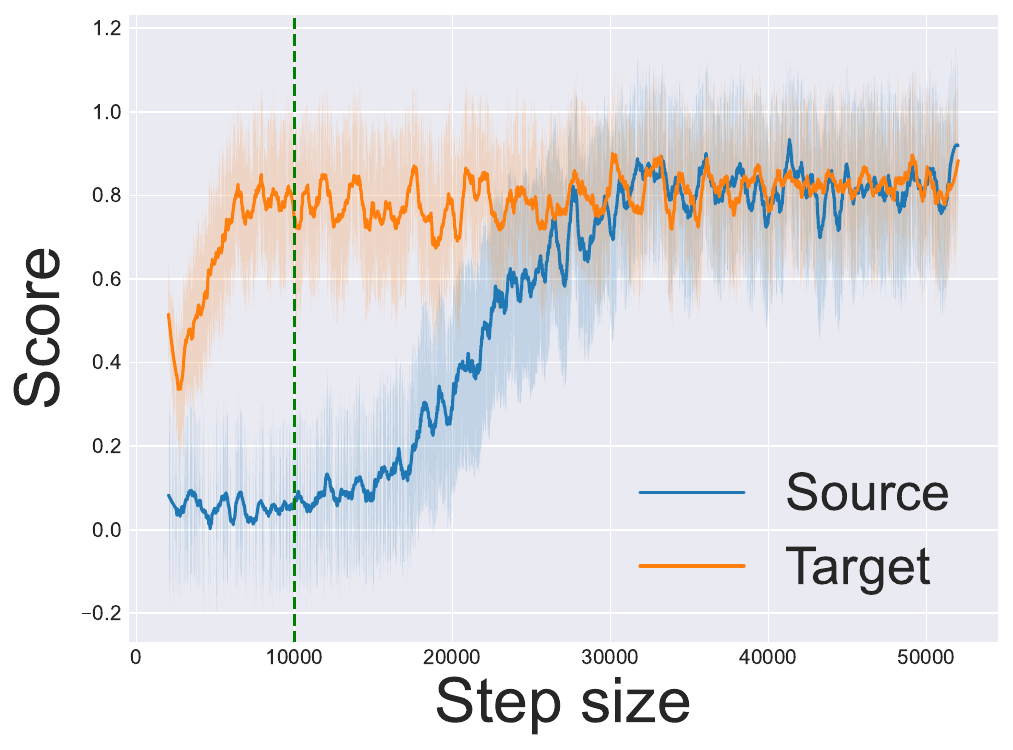}
        \caption{Fine-tuning in Target 2.}
        \label{fig:tr2}
    \end{subfigure}
    \hfill
    \begin{subfigure}[b]{0.3\linewidth}
        \centering
        \includegraphics[width=\linewidth]{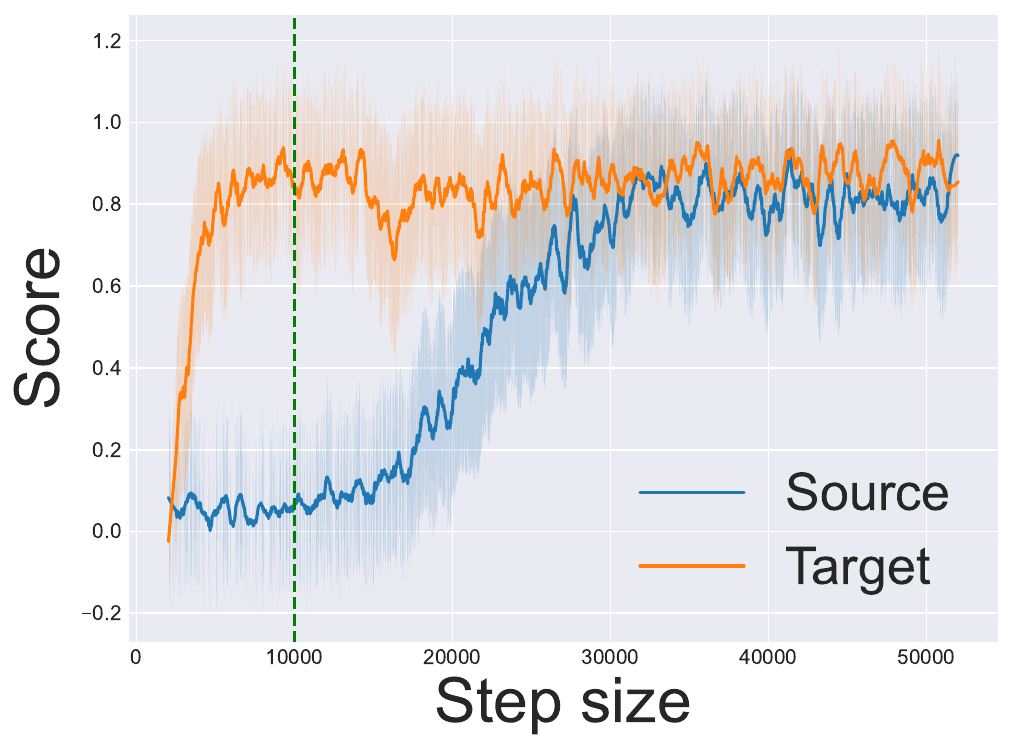}
        \caption{Fine-tuning in Target 3.}
        \label{fig:tr3}
    \end{subfigure}
    \caption{Figure~\ref{fig:minigrid_transfer} illustrates the reward curves during source task training and subsequent fine-tuning on target tasks in the MiniGrid environment. Figures~\ref{fig:tr1},~\ref{fig:tr2}, and~\ref{fig:tr3} correspond to Target tasks 1, 2, and 3, respectively. In Figure~\ref{fig:tr2}, the reward does not start from zero, as the agent already knows how to pick the green ball from the source task. As a result, it initially receives a partial reward (approximately 0.5) before learning to pick the red ball as required by the new task. The vertical green line indicates the minimum number of additional interactions needed for successful adaptation. All reward curves are smoothed and averaged over five random seeds.}
    \label{fig:minigrid_transfer}
    \vspace{15pt}
\end{figure}

\begin{figure}[htb]
    \centering
    \captionsetup[subfigure]{font=small, skip=0pt} 
    \captionsetup{skip=10pt} 
    \begin{subfigure}[b]{0.3\linewidth}
        \centering
        \includegraphics[width=\linewidth]{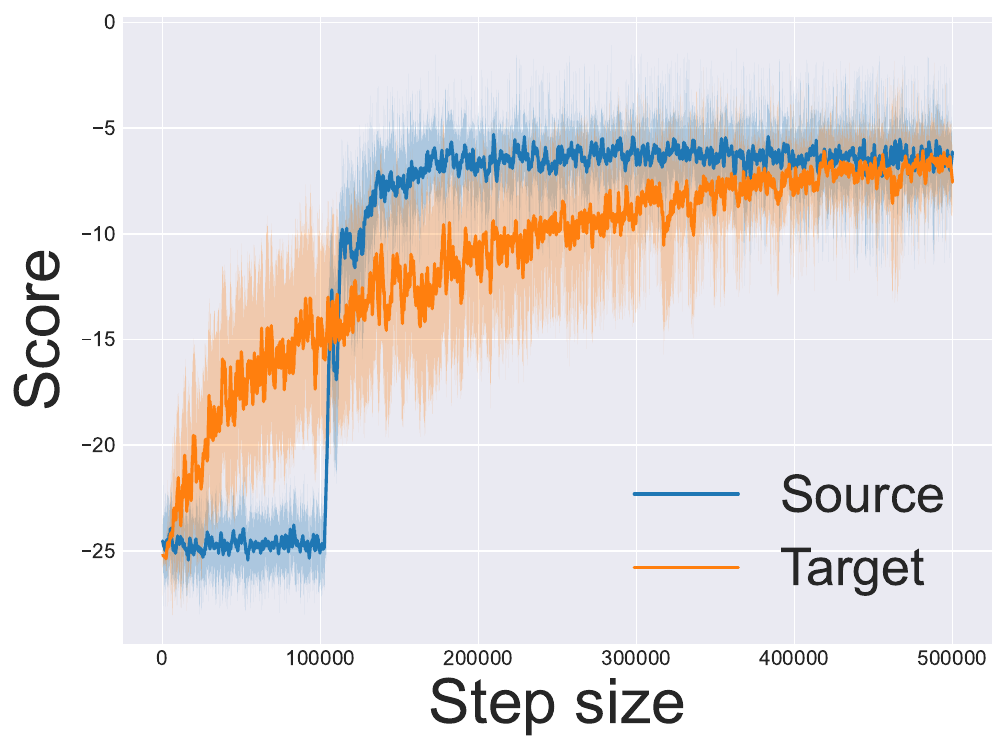}
        \caption{Fine-tuning for red target.}
        \label{fig:tr_reacher1}
    \end{subfigure}
    \hfill
    \begin{subfigure}[b]{0.3\linewidth}
        \centering
        \includegraphics[width=\linewidth]{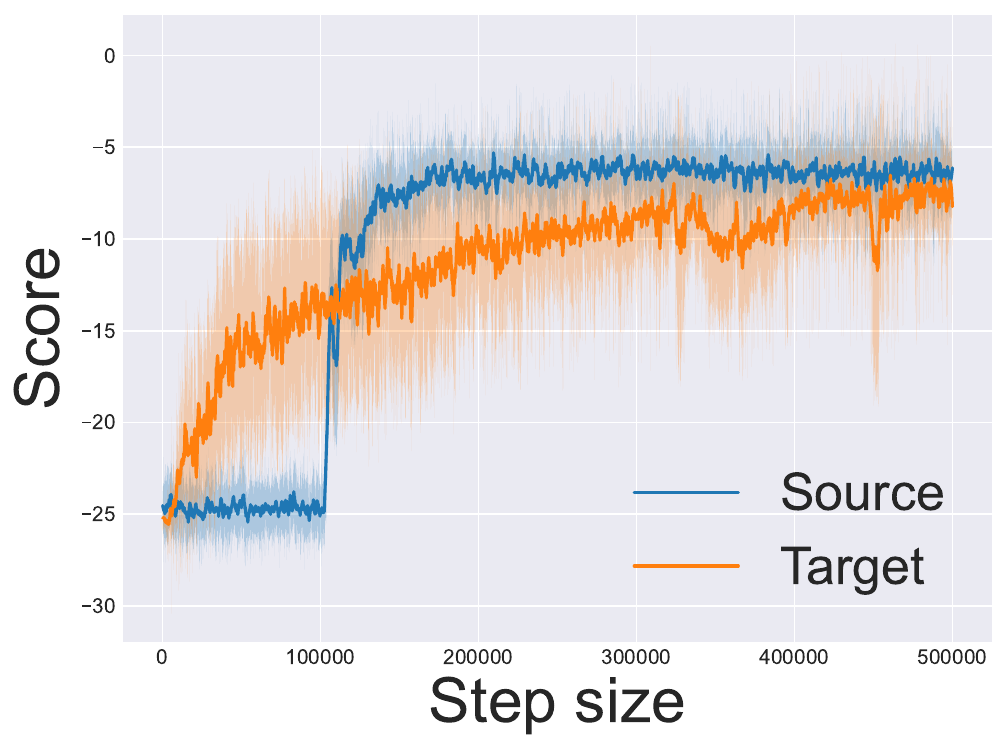}
        \caption{Fine-tuning for green target.}
        \label{fig:tr_reacher2}
    \end{subfigure}
    \hfill
    \begin{subfigure}[b]{0.3\linewidth}
        \centering
        \includegraphics[width=\linewidth]{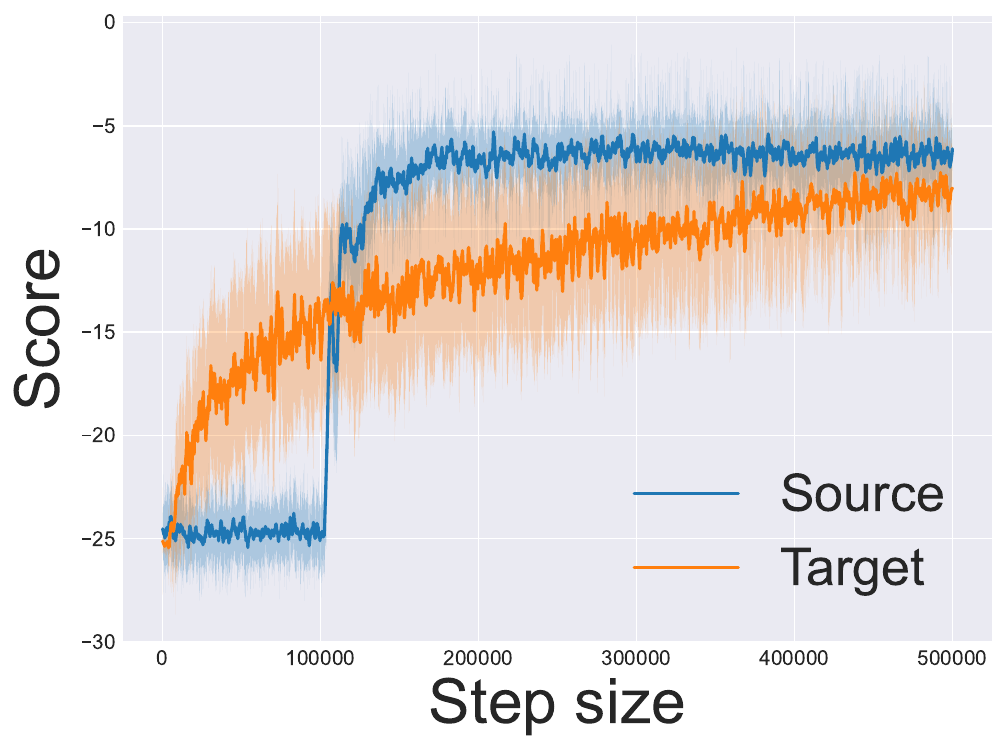}
        \caption{Fine-tuning for yellow target.}
        \label{fig:tr_reacher3}
    \end{subfigure}
    \caption{Figure~\ref{fig:reacher_transfer} shows the reward curves for both the source task training and the fine-tuning phase in the target tasks within the Reacher environment. Specifically, Figures~\ref{fig:tr_reacher1},~\ref{fig:tr_reacher2}, and~\ref{fig:tr_reacher3} correspond to the Red, Green, and Yellow target tasks, respectively. During source training, the agent explores randomly for the first 100K timesteps, which accounts for the sharp increase in reward observed after this point. In contrast, for the target tasks, training begins immediately from timestep zero. All reward curves are smoothed and averaged over five random seeds for clarity.}
    \label{fig:reacher_transfer}
    \vspace{15pt}
\end{figure}

\subsubsection{Evaluating \textbf{H2}: Data Efficiency}
\label{sec:data_efficiency}

We evaluate the data efficiency of our approach by comparing it to standard fine-tuning using Soft Actor-Critic (SAC). For fine-tuning, the SAC agent is initialized with weights learned in the source environment and allowed to interact with the target environment under the new reward function.

Our findings are as follows:
\begin{itemize}
    \item In the \textbf{MiniGrid} environment, fine-tuning requires approximately 10K additional interactions in the target environment with a modified reward function, as shown in Figure~\ref{fig:minigrid_transfer}. In contrast, in the \textbf{Reacher} environment, fine-tuning requires roughly the same number of interactions as were used during source training, as evidenced in Figure~\ref{fig:reacher_transfer}.
    
    \item \textit{MAGIK} achieves comparable performance using only \textbf{600 human-labeled examples}\footnote{We conducted additional experiments with only 120 data points, yielding promising results as presented in Supplementary Section~\ref{sec:add_exp}. However, we make a conservative claim in the main text.} in MiniGrid and \textbf{250 in Reacher}, representing less than \textbf{1\%} of the source training data.
\end{itemize}

Thus, \textit{MAGIK} provides up to a \textbf{16× reduction} (10K/600) in additional data requirements in each task in MiniGrid and performs even more efficiently in Reacher. Beyond data efficiency, \textit{MAGIK} also enables \textbf{zero-shot adaptation}, allowing agents to transfer knowledge without any further interaction and training for each task in the target environment. This demonstrates significantly improved \textbf{transferability} to novel but structurally similar tasks, while reducing the burden of environment-specific fine-tuning, thereby establishing \textbf{H2}.


\begin{figure}[h]
\centering
\includegraphics[width=\linewidth]{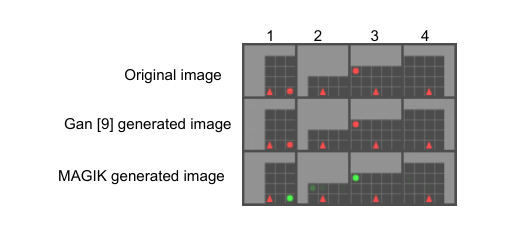}
\caption{Comparison of imagined states from a GAN-based method~\cite{gamrian2019transfer} and \textit{MAGIK}. The first row shows target domain observations, the second row depicts GAN-generated source states, and the third shows \textit{MAGIK}'s outputs conditioned to green class. MAGIK succeeds in converting red to green balls, though ambiguity appears when class features are absent because the model doesn't know where to place the objects as class objects are absent in the original image as in second and fourth column.}
\label{fig:gan}
\vspace{15pt}
\end{figure}

\subsection{Comparison with prior methods}
In this section, we analyse the applicability of prior methods \cite{gamrian2019transfer,higgins2017darla,zhu2024distributional,zhang2017,touati2022does,chua2024learning} to our target tasks. The approach in \cite{gamrian2019transfer} relies on domain adaptation and assumes that the source and target observations come from distinct distributions. However, this assumption does not hold in Targets 2 and 3, where the source and target environments are identical. Therefore, we evaluate this method only on Target 1, and the results are presented in Table~\ref{tab:transfer}. Notably, \cite{gamrian2019transfer} requires explicit access to both source and target environments, while our method does not. Nevertheless, for a fair comparison, we allow the agent to interact with both environments during evaluation.

The failure of \cite{gamrian2019transfer} in this setting can be attributed to the structure of the source and target distributions. In Target 1, the target distribution is a strict subset of the source—while the source environment includes both red and green objects, the target environment contains only red objects. As a result, when the GAN attempts to map target observations into the source domain, it often leaves them unchanged, since red-only views already exist in the source distribution. This causes the GAN to fail at transforming red objects into green ones as expected. This limitation is visually evident in Figure~\ref{fig:gan}, where the GAN-generated target observations remain unaltered.

The method proposed in \cite{higgins2017darla} is designed for generalisation across visual changes in the background, assuming the reward function remains constant. For example, an agent trained to pick oranges in a blue room should be able to generalise to picking oranges in a red room. However, in our setting, the reward function itself differs between the source and target tasks, rendering direct application of this method infeasible.

We also evaluate methods based on variants of Successor Features (SF) \cite{zhang2017,touati2022does,chua2024learning}. While these methods aim to facilitate transfer through decoupling dynamics and rewards, we found that SF-based approaches struggle to match the adaptation capability of our proposed MAGIK method in both environments. The comparison includes fine-tuning the SF in the target environment as well as direct transfer via linear regression on the reward function (More details are available in supplementary Section \ref{sec:sf}). In both cases, the agent fails to complete the task successfully, as shown in Tables~\ref{tab:transfer} and~\ref{tab:transfer_reacher}. Even with explicit deployment in the target domain—15K timesteps in MiniGrid and 1M in Reacher—SF methods do not learn sufficiently generalisable representations for seamless transfer.

Lastly, we do not compare against the full method in \cite{zhu2024distributional}, as it employs a diffusion model for imagination, which we consider unnecessary given the simplicity of the environments studied. Instead, we adapt the general idea of SF from their work and compare this simplified version against our method.

\section{Discussion and Conclusion}
\label{sec:discussion_conclusion}

We introduced \textbf{MAGIK}, a generative framework for transfer learning in Deep RL that enables agents to adapt zero-shot to unseen but analogous tasks by imagining source-aligned states from target observations. By disentangling task-agnostic and class-specific information using a semi-supervised VAE, MAGIK allows controlled synthesis of observations that align with prior experience, facilitating policy reuse without additional training in the target task.

Our empirical evaluations show that MAGIK outperforms conventional transfer approaches, including domain adaptation \cite{gamrian2019transfer} and successor feature-based methods \cite{zhang2017,touati2022does,chua2024learning}, especially when source and target environments share structural similarity but differ in task semantics (e.g., goals). Unlike methods requiring access to the target domain or complex training regimes, MAGIK leverages a lightweight imagination module trained solely on offline data from the source task.

Despite these strengths, MAGIK relies on the assumption that each observation can be cleanly decomposed into disentangled latent variables: class-agnostic (\(z\)) and class-specific (\(c\)) components. In real-world settings, this assumption may not always hold due to entangled or ambiguous observations. To mitigate this, future work could explore relaxed forms of disentanglement (e.g., partial or soft factorisation), self-supervised or contrastive objectives to reduce reliance on labels, and attention-based or probabilistic imagination networks to improve robustness under uncertainty. Overall, MAGIK offers a scalable and conceptually simple foundation for imagination-driven policy transfer in RL.









\bibliography{mybibfile}
\newpage
\onecolumn
\appendix
\section{Derivation of ELBO}
\label{sec:elbo}

As discussed in Section~\ref{sec:im_training}, each observation is assumed to be explained by two complementary latent variables: a continuous latent \(z\) that captures task-agnostic factors and a discrete latent \(c \in \{1, \dots, K\}\) that captures task-specific factors. Together, these variables are designed to be mutually exclusive (each explains a distinct source of variation) and collectively sufficient to reconstruct the observation. 

We now formalise this intuition within a probabilistic framework. 
The generative model for \(x\) is parameterized by \(\boldsymbol{\theta}\), while the variational distributions are parameterized by \(\boldsymbol{\phi}\). 
The priors and the label model \(p(y \mid c)\) are fixed and unparameterized. 
Formally, the joint model is given by:
\[
p(x, y, z, c) = p_{\boldsymbol{\theta}}(x \mid z, c)\,p(y \mid c)\,p(z)\,p(c),
\]
with the variational approximation factorised as:
\[
q_{\boldsymbol{\phi}}(z, c \mid x) = q_{\boldsymbol{\phi}}(z \mid x)\,q_{\boldsymbol{\phi}}(c \mid x).
\]

For completeness, a step-by-step derivation of the ELBO for both labelled and unlabelled cases is provided below.

\subsection{Labelled Case (\(x, y\) known)}

We aim to maximize the marginal likelihood:
\[
\log p(x, y) = \log \int_z \sum_{c=1}^K p_{\boldsymbol{\theta}}(x \mid z, c)\,p(y \mid c)\,p(z)\,p(c)\,dz.
\]

We introduce the variational approximation \(q_{\boldsymbol{\phi}}(z, c \mid x) = q_{\boldsymbol{\phi}}(z \mid x)\,q_{\boldsymbol{\phi}}(c \mid x)\) and apply Jensen’s inequality:
\[
\begin{aligned}
\log p(x, y)
&= \log \int_z \sum_{c=1}^K q_{\boldsymbol{\phi}}(z \mid x)\,q_{\boldsymbol{\phi}}(c \mid x)\,
\frac{p_{\boldsymbol{\theta}}(x \mid z, c)\,p(y \mid c)\,p(z)\,p(c)}{q_{\boldsymbol{\phi}}(z \mid x)\,q_{\boldsymbol{\phi}}(c \mid x)}\,dz \\
&\geq \sum_{c=1}^K q_{\boldsymbol{\phi}}(c \mid x)\int_z q_{\boldsymbol{\phi}}(z \mid x)
\log \left( \frac{p_{\boldsymbol{\theta}}(x \mid z, c)\,p(y \mid c)\,p(z)\,p(c)}{q_{\boldsymbol{\phi}}(z \mid x)\,q_{\boldsymbol{\phi}}(c \mid x)} \right) dz.
\end{aligned}
\]

This gives the labelled ELBO:
\[
\begin{aligned}
\mathcal{L}_{\text{label}}(x, y; \boldsymbol{\theta}, \boldsymbol{\phi})
&= \sum_{c=1}^K q_{\boldsymbol{\phi}}(c \mid x)\,\mathbb{E}_{q_{\boldsymbol{\phi}}(z \mid x)}\left[ \log p_{\boldsymbol{\theta}}(x \mid z, c) \right] \\
&\quad + \sum_{c=1}^K q_{\boldsymbol{\phi}}(c \mid x)\,\log p(y \mid c) \\
&\quad - \mathrm{KL}(q_{\boldsymbol{\phi}}(z \mid x) \parallel p(z)) 
- \mathrm{KL}(q_{\boldsymbol{\phi}}(c \mid x) \parallel p(c)).
\end{aligned}
\]

\paragraph{Supervision Term.}
The term
\[
\sum_{c=1}^K q_{\boldsymbol{\phi}}(c \mid x)\,\log p(y \mid c)
\]
encourages the variational posterior over classes \(q_{\boldsymbol{\phi}}(c \mid x)\) to concentrate on the true label \(y\).  

We typically define \(p(y \mid c)\) as a deterministic distribution:
\[
p(y \mid c) = \delta_{c = y},
\]
where \(\delta_{c=y} = 1\) if \(c = y\), and 0 otherwise. Consequently,
\[
\log p(y \mid c) =
\begin{cases}
0 & \text{if } c = y, \\
-\infty & \text{otherwise}.
\end{cases}
\]

Thus, if \(q_{\boldsymbol{\phi}}(c \mid x)\) assigns non-zero probability to any class \(c \neq y\), the corresponding term contributes 
\(q_{\boldsymbol{\phi}}(c \mid x)(-\infty)\), which collapses the ELBO to \(-\infty\). 
Maximization of the ELBO therefore forces the model to place almost all mass on the correct class. 
Formally, this reduces to:
\[
\sum_{c=1}^K q_{\boldsymbol{\phi}}(c \mid x)\,\log p(y \mid c) 
= \log q_{\boldsymbol{\phi}}(c = y \mid x).
\]

This is exactly the log-likelihood of the correct class under the variational posterior, 
equivalent to the negative cross-entropy loss used in supervised classification.

\paragraph{Interpretation.}
This derivation shows that the supervision term not only promotes the correct label but also 
heavily penalises probability mass on incorrect labels. 
In doing so, it explicitly supervises the inference network \(q_{\boldsymbol{\phi}}(c \mid x)\) to predict the semantic label \(y\), 
rather than learning \(c\) through unsupervised clustering, which may not align with actual class semantics.

\paragraph{Full Labelled ELBO.}
\[
\boxed{
\begin{aligned}
\mathcal{L}_{\text{label}}(x, y; \boldsymbol{\theta}, \boldsymbol{\phi})
&= \underbrace{\sum_{c} q_{\boldsymbol{\phi}}(c \mid x)\,\mathbb{E}_{q_{\boldsymbol{\phi}}(z \mid x)}\left[ \log p_{\boldsymbol{\theta}}(x \mid z, c) \right]}_{\text{(A) Reconstruction}} \\[6pt]
&\quad + \underbrace{\log q_{\boldsymbol{\phi}}(c = y \mid x)}_{\text{(B) Supervision (cross-entropy)}} \\[6pt]
&\quad - \underbrace{\mathrm{KL}(q_{\boldsymbol{\phi}}(z \mid x) \parallel p(z))}_{\text{(C) KL for } z}
\;\; - \underbrace{\mathrm{KL}(q_{\boldsymbol{\phi}}(c \mid x) \parallel p(c))}_{\text{(D) KL for } c}.
\end{aligned}
}
\]

Here, term (B) corresponds exactly to the supervised cross-entropy objective, 
with incorrect class assignments effectively penalized with infinite cost in the ELBO formulation.

\subsection{Unlabelled Case (only \(x\) known)}

When \(y\) is unobserved, the marginal log-likelihood becomes:
\[
\log p(x) = \log \int_z \sum_{c=1}^K p_{\boldsymbol{\theta}}(x \mid z, c)\,p(z)\,p(c)\,dz.
\]

Again, applying variational inference:
\[
\begin{aligned}
\log p(x) 
&= \log \int_z \sum_{c=1}^K q_{\boldsymbol{\phi}}(z \mid x)\,q_{\boldsymbol{\phi}}(c \mid x)\,
\frac{p_{\boldsymbol{\theta}}(x \mid z, c)\,p(z)\,p(c)}{q_{\boldsymbol{\phi}}(z \mid x)\,q_{\boldsymbol{\phi}}(c \mid x)}\,dz \\
&\geq \sum_{c=1}^K q_{\boldsymbol{\phi}}(c \mid x) \int_z q_{\boldsymbol{\phi}}(z \mid x)
\log \left( \frac{p_{\boldsymbol{\theta}}(x \mid z, c)\,p(z)\,p(c)}{q_{\boldsymbol{\phi}}(z \mid x)\,q_{\boldsymbol{\phi}}(c \mid x)} \right) dz.
\end{aligned}
\]

This gives the ELBO for unlabelled data:
\[
\begin{aligned}
\mathcal{L}_{\text{unlabel}}(x; \boldsymbol{\theta}, \boldsymbol{\phi})
&= \sum_{c=1}^K q_{\boldsymbol{\phi}}(c \mid x)\,\mathbb{E}_{q_{\boldsymbol{\phi}}(z \mid x)}\left[ \log p_{\boldsymbol{\theta}}(x \mid z, c) \right] \\
&\quad - \mathrm{KL}(q_{\boldsymbol{\phi}}(z \mid x) \parallel p(z))
- \mathrm{KL}(q_{\boldsymbol{\phi}}(c \mid x) \parallel p(c)).
\end{aligned}
\]

\subsection{Total Objective}

The full training objective over a labelled dataset \(\mathcal{D}_\ell = \{(x_i, y_i)\}\) and unlabelled dataset \(\mathcal{D}_u = \{x_j\}\) is:
\[
\mathcal{L}_{\text{total}}(\boldsymbol{\theta}, \boldsymbol{\phi}) =
\sum_{(x,y) \in \mathcal{D}_\ell} \mathcal{L}_{\text{label}}(x, y; \boldsymbol{\theta}, \boldsymbol{\phi})
+ \sum_{x \in \mathcal{D}_u} \mathcal{L}_{\text{unlabel}}(x; \boldsymbol{\theta}, \boldsymbol{\phi}).
\]

\section{Disentanglement evaluation}
\label{sec:disentanglement}
We assume that $z$ and $c$ are independent and enforce this criterion through an additional Hilbert-Schmidt Independence Criterion (HSIC) auxiliary loss. To evaluate the disentanglement, we traverse one latent variable while keeping the other fixed to observe whether the decoder generates meaningful outputs.

\begin{figure}[h]
    \begin{center}
        \includegraphics[width=\linewidth]{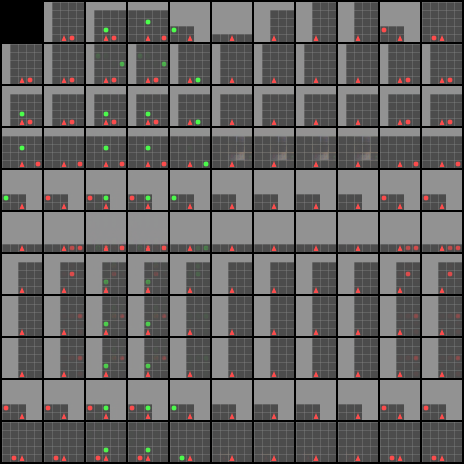}
    \end{center}
    \caption{Image generated by traversing through $z$ and $c$.}
    \label{fig:traversal}
    \vspace{15pt}
\end{figure}

Figure~\ref{fig:traversal} presents images generated in a structured manner: class-agnostic attributes are derived from the first column, while class-related features originate from the first row. Specifically, the image at position $(i,j)$ combines the class-agnostic features of the $i^{th}$ image in the first column with the class-related attributes of the $j^{th}$ image in the first row. 

The traversal results demonstrate that class-agnostic features, such as walls and empty spaces, are preserved horizontally across each row. Interestingly, the latent variable $z$ not only captures class-agnostic scene elements but also determines object placement, while $c$ solely governs the object identity. This is evident in the third and fourth columns, where the model removes objects that do not belong to the given class label. Additionally, when the reference images contain no objects, the model struggles with object placement, leading to artefacts in the generated images—particularly noticeable in the $6^{th}$ to $8^{th}$ rows. 

These observations indicate that the model successfully disentangles the class-agnostic latent variable ($z$) from the class-specific latent variable ($c$).

\section{Network Architecture}
\label{sec:detailed_archtecture}
We detail the network architectures used for different environments. For image-based observations in MiniGrid, we use a CNN-based encoder and decoder. For the Reacher environment, where the observations are feature-based, we employ MLP-based encoder and decoder architectures.

\textbf{Reacher (MLP Encoder-Decoder).}  
The MLP encoder consists of two linear layers with hidden dimensions of 128, with a BatchNorm layer between them and a final ReLU activation. This is followed by a residual block, and finally a linear layer that maps to a 32-dimensional encoder output. The output from the class-agnostic encoder is passed through a bottleneck layer with a latent dimension \(z\) of size 11. The class-specific encoder maps the features to a \(c\) latent space of dimension 4, corresponding to the number of classes. The decoder is a three-layer MLP with hidden dimensions 256 and 128. FiLM layers are added between each hidden layer to condition on the \(z\) and \(c\) latents.

\textbf{MiniGrid (CNN Encoder-Decoder).}  
The CNN encoder details are provided in Table~\ref{tab:encoder_arch}. After the convolutional layers, we apply two residual blocks followed by an MLP with hidden dimensions 2048 and 512. The decoder first maps the \(z\)-latent to a shape of \((256, 10, 10)\), matching the final output size of the CNN encoder. This output is reshaped and passed through a residual block, followed by a FiLM layer conditioned on the \(c\)-latent. Two successive upsampling layers are used to gradually increase the spatial resolution to \(40 \times 40\), while reducing the channel dimension to 3. Finally, another FiLM layer is applied, followed by a Sigmoid activation to generate the reconstructed image.

\begin{table}[h]
\caption{\label{tab:encoder_arch} CNN encoder architecture details.}
\centering
\begin{tabular}{|l|l|}
\hline
\textbf{Particulars} & \textbf{Values} \\ \hline
Channels & 64, 128, 256 \\ \hline
Kernel Size & 7, 4, 4 \\ \hline
Stride & 1, 2, 2 \\ \hline
Padding & 3, 1, 1 \\ \hline
\end{tabular}
\end{table}

\section{Training hyper-parameters}
The hyper-parameters used  while training are given in table Table \ref{tab:optimizer_vae}.
\begin{table}
\caption{\label{tab:optimizer_vae} Table showing the VAE Optimizer details}
\centering
\begin{tabular}{|l|l|} 
 \hline
  Particulars & Value\\ [0.5ex] 
 \hline
 Learning rate& $5e-4$\\ 
\hline
 Batch size& $100$\\
 \hline
 KL divergence weight& $0.01$\\ [1ex] 
 \hline
 HSIC weight& $0.10$\\ [1ex] 
 \hline
 Reconstruction loss weight& $2$\\ [1ex] 
 \hline
 label loss weight& $5$\\ [1ex] 
 \hline
 Gradient clipping (Max norm)& $1$\\ [1ex] 
 \hline
 Optimizer& ADAM\\ [1ex] 
 \hline
\end{tabular}
\end{table}

\section{Successor Feature Training}
\label{sec:sf}
\begin{figure}[h]
    \centering
    \captionsetup[subfigure]{font=small, skip=5pt}
    \captionsetup{skip=12pt}
    \begin{subfigure}[b]{0.45\linewidth}
        \centering
        \includegraphics[width=\linewidth]{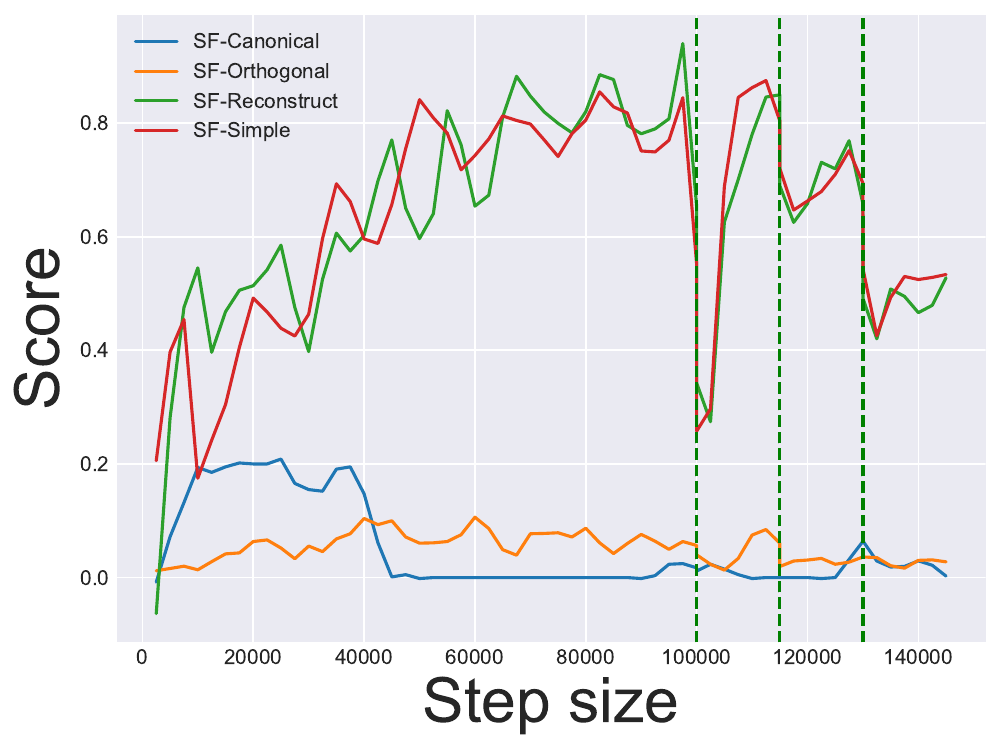}
        \caption{SF evaluation during training in MiniGrid.}
        \label{fig:sf_fine_tuning1}
    \end{subfigure}
    \hfill
    \begin{subfigure}[b]{0.45\linewidth}
        \centering
        \includegraphics[width=\linewidth]{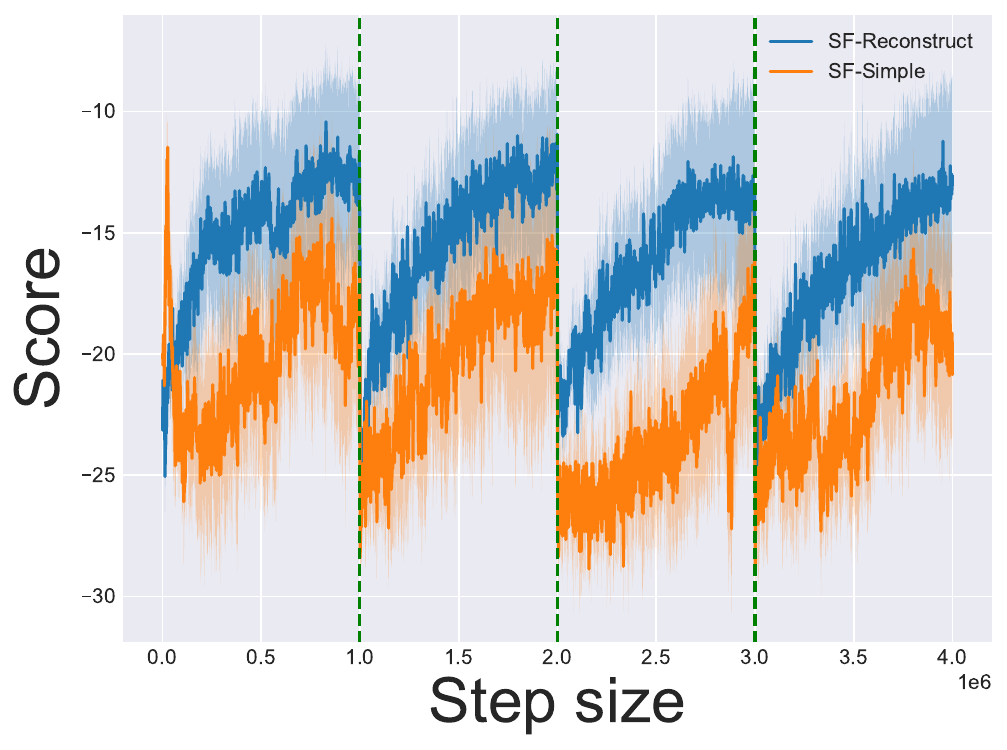}
        \caption{SF evaluation during training in Reacher.}
        \label{fig:sf_fine_tuning2}
    \end{subfigure}
    \caption{Agent evaluation at regular intervals during SF training. The vertical green dashed line indicates the point of task change. The replay buffer is reset when the task changes.}
    \label{fig:sf_fine_transfer}
\end{figure}

We fine-tune the Successor Features (SFs) learned in the source task by collecting state-action pairs and their corresponding rewards in the target task. However, we observe that even the fine-tuning of SFs is not sufficient to match the performance achieved by \textit{MAGIK}. The learning curve for the fine-tuning of the SF is given in Figure \ref{fig:sf_fine_transfer}

We also evaluate the direct transfer of SFs by performing linear regression between the collected $\phi(s, a)$ features and the rewards $r(s, a)$ in the target task to estimate the reward weight vector $w$. However, results suggest that this approach fails to solve the task, with the agent unable to succeed in any rollout instance.

\section{Imagination using GAN}
\label{sec:gan_imagination}

This section presents the framework of Generative Adversarial Network (GAN) alternative of the Imagination network proposed in the main paper, which integrates conditional generation, semi-supervised learning, and disentanglement via gradient reversal. The model comprises three key components: the Discriminator, the Auxiliary Classifier, and the Encoder-Decoder pair. Each component is trained with specific loss functions to achieve accurate classification, realistic image generation, and class-agnostic latent representations.

The Discriminator \( D \) classifies input images into one of \( K+1 \) classes, where the first \( K \) classes represent real class labels, and the additional class denotes fake images. Its loss combines three terms. For labeled real images \( x^{(l)} \) with labels \( y^{(l)} \), the cross-entropy loss encourages correct classification, where \( P_D(k \mid x) = \frac{\exp(D(x)_k)}{\sum_{m=0}^{K-1} \exp(D(x)_m)} \), and \( N_l \) is the number of labeled samples:

\begin{eqnarray}
L_D^{\text{labeled}} &=& -\frac{1}{N_l} \sum_{i=1}^{N_l} \log P_D(y^{(l)}_i \mid x^{(l)}_i)
\end{eqnarray}

For real images \( x^{(r)} \) (labeled and unlabeled), with \( N_r \) as the number of real images and \( P_D(\text{fake} \mid x) = \text{softmax}(D(x))_{\text{fake}} \), the Discriminator ensures low fake-class probability:

\begin{eqnarray}
L_D^{\text{unlabeled}} &=& -\frac{1}{N_r} \sum_{i=1}^{N_r} \log \left(1 - P_D(\text{fake} \mid x^{(r)}_i)\right)
\end{eqnarray}

For fake images \( x^{(f)} \), with \( N_f \) as the number of fake images, the Discriminator maximizes fake-class probability:

\begin{eqnarray}
L_D^{\text{fake}} &=& -\frac{1}{N_f} \sum_{i=1}^{N_f} \log P_D(\text{fake} \mid x^{(f)}_i)
\end{eqnarray}

The total Discriminator loss is:

\begin{eqnarray}
L_D &=& L_D^{\text{labeled}} + L_D^{\text{unlabeled}} + L_D^{\text{fake}}
\end{eqnarray}

In expectation, this is:

\begin{eqnarray}
L_D &=& \mathbb{E}_{x^{(l)}, y^{(l)}} \left[-\log P_D(y^{(l)} \mid x^{(l)})\right] + \mathbb{E}_{x^{(r)}} \left[-\log \left(1 - P_D(\text{fake} \mid x^{(r)})\right)\right] + \mathbb{E}_{x^{(f)}} \left[-\log P_D(\text{fake} \mid x^{(f)})\right]
\end{eqnarray}

The Auxiliary Classifier \( C \) predicts class labels from class-agnostic latent representations \( z = E(x^{(l)}) \), where \( E \) is the Encoder and \( P_C(k \mid z) = \text{softmax}(C(z))_k \). It is trained on labeled data:

\begin{eqnarray}
L_C &=& -\frac{1}{N_l} \sum_{i=1}^{N_l} \log P_C(y^{(l)}_i \mid z_i)
\end{eqnarray}

Here, \( z \) is detached to prevent gradients from affecting \( E \), updating \( C \) to improve class prediction.

The Encoder \( E \) and Decoder \( G \) are trained jointly with reconstruction, adversarial, and disentanglement losses. The reconstruction loss ensures accurate image reconstruction. For labeled data:

\begin{eqnarray}
L_R^{\text{labeled}} &=& \frac{1}{N_l} \sum_{i=1}^{N_l} \left\| G(E(x^{(l)}_i), y^{(l)}_i) - x^{(l)}_i \right\|_2^2
\end{eqnarray}

For unlabeled data, using pseudo-labels \( \hat{y}_u = \arg\max_{k=0,\ldots,K-1} D(x^{(u)})[:, :-1] \):

\begin{eqnarray}
L_R^{\text{unlabeled}} &=& \frac{1}{N_u} \sum_{i=1}^{N_u} \left\| G(E(x^{(u)}_i), \hat{y}_{u,i}) - x^{(u)}_i \right\|_2^2
\end{eqnarray}

The total reconstruction loss is:

\begin{eqnarray}
L_R &=& L_R^{\text{labeled}} + L_R^{\text{unlabeled}}
\end{eqnarray}

The adversarial loss encourages realistic fake images \( x^{(f)} = G(E(x), y') \), with random label \( y' \):

\begin{eqnarray}
L_A &=& -\frac{1}{N_r} \sum_{i=1}^{N_r} \log \left(1 - P_D(\text{fake} \mid x^{(f)}_i)\right)
\end{eqnarray}

The disentanglement loss uses a gradient reversal layer (GRL) to make \( z \) class-agnostic:

\begin{eqnarray}
L_{\text{dis}} &=& -\frac{1}{N_l} \sum_{i=1}^{N_l} \log P_C(y^{(l)}_i \mid z_{\text{reversed},i})
\end{eqnarray}

Here, \( z_{\text{reversed}} = \text{GRL}(E(x^{(l)})) \), and the gradient is scaled by \( -\lambda_{\text{dis}} \). The total Encoder-Decoder loss is:

\begin{eqnarray}
L_{EG} &=& L_R + L_A + \lambda_{\text{dis}} \cdot L_{\text{dis}}
\end{eqnarray}

Encoder gradients are:

\begin{eqnarray}
\frac{\partial L_{EG}}{\partial E} &=& \frac{\partial L_R}{\partial E} + \frac{\partial L_A}{\partial E} - \lambda_{\text{dis}} \cdot \frac{\partial L_{\text{dis}}}{\partial E}
\end{eqnarray}

This promotes class-agnostic representations, while the Decoder minimises only reconstruction and adversarial losses.

In summary, the proposed GAN alternative of the Imagination network combines conditional generation, semi-supervised learning via pseudo-labels, and disentanglement through gradient reversal. The Discriminator ensures accurate classification, the Auxiliary Classifier refines class predictions, and the Encoder-Decoder pair reconstructs images, generates realistic samples, and learns disentangled representations, making the model a robust framework for conditional image generation.

\section{Additional Experiment}
\label{sec:add_exp}
The results detailed in Table~\ref{tab:transfer_reacher} were obtained using a diverse dataset, where 100K data points were collected purely from random exploration, in the full SAC training data. However, we observed a decline in performance when the source training data lacked sufficient diversity. Specifically, we tested MAGIK's performance using SAC training data with only 30K and 50K random exploration samples. The results of this experiment are provided in Table~\ref{tab:decline_result}.

\begin{table}[htbp]
\caption{Performance of MAGIK with different levels of data diversity. Results are averaged over 5 seeds, each with 20 evaluation episodes.}
\label{tab:decline_result}
\centering
\begin{tabular}{lll}
\toprule
\textbf{Target} & \textbf{Random Samples} & \textbf{Target Reached (Out of 20)} \\
\midrule
\multirow{2}{*}{Red}    & 5K  & $13.00 \pm 1.22$ \\
                        & 10K & $15.00 \pm 1.58$ \\
                        & 30K & $16.40 \pm 1.51$ \\
                        & 50K & $15.40 \pm 1.51$ \\
\midrule
\multirow{2}{*}{Green}  & 5K  & $12.80 \pm 1.30$ \\
                        & 10K & $14.00 \pm 2.54$ \\
                        & 30K & $15.40 \pm 2.70$ \\
                        & 50K & $14.20 \pm 1.78$ \\
\midrule
\multirow{2}{*}{Blue}   & 30K & $20.00 \pm 0.00$ \\
                        & 50K & $20.00 \pm 0.00$ \\
                        & 30K & $20.00 \pm 0.00$ \\
                        & 50K & $20.00 \pm 0.00$ \\
\midrule
\multirow{2}{*}{Yellow} & 30K & $12.80 \pm 1.22$ \\
                        & 50K & $16.66 \pm 0.89$ \\
                        & 30K & $15.40 \pm 1.14$ \\
                        & 50K & $16.20 \pm 1.78$ \\
\bottomrule
\end{tabular}
\end{table}

To further evaluate data efficiency, we conducted experiments with an even smaller dataset of only 120 data points, as reported in the main paper. Remarkably, MAGIK still demonstrated promising performance, as shown in Table \ref{tab:transfer_MiniGrid_100}.

\begin{table}[htbp]
\caption{Performance of MAGIK in MiniGrid env with 100 labelled data.}
\label{tab:transfer_MiniGrid_100}
\centering
\begin{tabular}{lll}
\toprule
\multicolumn{1}{l}{\bf Target} & \multicolumn{1}{l}{\bf \makecell{Green ball picked \\ (Out of 20)}} & \multicolumn{1}{l}{\bf \makecell{Red ball picked \\ (Out of 20)}} \\
\midrule
\multirow{1}{*}{1}      & NA              & $9.00 \pm 0.70$   \\
\midrule
\multirow{1}{*}{2}      & $8.60 \pm 1.14$ & $8.80 \pm 0.83$   \\
\midrule
\multirow{1}{*}{3}      & $8.70 \pm 2.38$ & $0.60 \pm 1.34$  \\

\bottomrule
\end{tabular}
\end{table}

\end{document}